\documentclass[11pt,a4paper]{article}
\usepackage[hyperref]{emnlp2020}
\usepackage[normalem]{ulem}
\usepackage{booktabs}
\usepackage{times}
\usepackage{latexsym}

\usepackage{url}

\usepackage{multirow}
\usepackage{setspace}

\usepackage{tabularx}
\newcolumntype{L}{>{\raggedright\arraybackslash}X}

\usepackage{microtype}
\usepackage{graphicx}

\aclfinalcopy % Uncomment this line for the final submission
\usepackage{xcolor}

\title{\textsc{TweetEval}: \\ Unified Benchmark and Comparative Evaluation for Tweet Classification}

\author{
 Francesco Barbieri$^{\clubsuit}$ ~ Jose Camacho-Collados$^{\dagger}$ \\
 \textbf{Leonardo Neves}$^{\clubsuit}$ ~ \textbf{Luis Espinosa-Anke}$^{\dagger}$  \vspace{0.1cm} \\
$^\clubsuit$Snap Inc., Santa Monica, CA 90405, USA\\
$^\dagger$School of Computer Science and Informatics, Cardiff University, United Kingdom \vspace{0.1cm} \\
{ \tt $^{\clubsuit}$\{fbarbieri,lneves\}@snap.com},\\
{  \tt $^{\dagger}$\{camachocolladosj,espinosa-ankel\}@cardiff.ac.uk}\\ 
}

\date{}

\begin{document}
\maketitle
\begin{abstract}

%Noisy user-generated content is one of the most challenging domains in natural language processing. Text in social media is typically short, plays around platform-specific restrictions, and features an ever-evolving pool of slang, jargon, hashtags and memes. 
The experimental landscape in natural language processing for social media is too fragmented. Each year, new shared tasks and datasets are proposed, ranging from classics like sentiment analysis to irony detection or emoji prediction. 
%Today, in the age of large pretrained language models, 
Therefore, it is unclear what the current state of the art is%for the \red{type of text found in social media}
, as there is no standardized evaluation protocol, neither a strong set of baselines trained on such domain-specific data. In this paper,
%Applying standard NLP techniques in social media has always been challenging. One of the main issues that have prevented standard evaluation protocols is the lack of standard benchmarks such as SentEval or GLUE, which are only available for other types of text. With this goal in mind, in this paper, 
we propose a new evaluation framework (\textsc{TweetEval}) consisting of seven heterogeneous Twitter-specific classification tasks. %One of the main features of this benchmark is its simplicity for evaluating new models, as all tasks are framed as tweet classification. 
We also provide a strong set of baselines as starting point, and compare different language modeling pre-training strategies. Our initial experiments show the effectiveness of starting off with existing pre-trained generic language models, and continue training them on Twitter corpora. 

\end{abstract}

%\begin{spacing}{0.99}

\section{Introduction}

Modern NLP systems are typically ill-equipped when applied to noisy user-generated text. The high-paced, conversational and idiosyncratic nature of social media, paired with platform-specific restrictions (e.g., Twitter's character limit), requires tackling additional challenges, for example, POS tagging \cite{derczynski-etal-2013-twitter}, lexical normalization \cite{han2011lexical,baldwin2015shared}, or named entity recognition \cite{ritter2011named,baldwin-etal-2013-noisy}. In other more generic contexts, these challenges can be considered solved or are simply non-existent. Moreover, other apparently simple tasks such as sentiment analysis have proven to be hard on Twitter data \cite{poria2020beneath}, among others, due to limited amount of contextual cues available in short texts \cite{kim2014sociolinguistic}.
%Moreover, the limited length restricts the presence of contextual cues normally present in dialogues or documents
%Moreover, other apparently simple tasks such as language identification have proven to be hard on short text \cite{lui2014accurate}, among others, due to frequent code-switching on social media \cite{kim2014sociolinguistic}.
In addition to these and other inherent difficulties, advances in NLP for user-generated data are hindered by its highly fragmented landscape and the lack of a unified evaluation framework. In the current era of pretraining and Language Models (LMs), this is particularly relevant, as these models exhibit a versatility that currently cannot be gauged comparably across Twitter datasets and tasks. This is not the case, however, in more ordinary textual genres and domains. For instance, well known benchmarks like SentEval \cite{conneau-kiela-2018-senteval}, GLUE \cite{wang2019glue} and SuperGLUE \cite{wang2019superglue} include standard NLP tasks such as language inference, paraphrase detection or sentiment analysis, among others. It is undisputable that these benchmarks have contributed to the fast development of language understanding techniques, and LMs in particular, as they have enabled comprehensive evaluations across several tasks in fair and reproducible conditions.

We thus take inspiration from the above to develop \textsc{TweetEval}, a benchmark for tweet classification in English. %, a tweet classification benchmark. 
\textsc{TweetEval} is a standardized test bed for seven tweet classification tasks. These are: sentiment analysis, emotion recognition, offensive language detection, hate speech detection, stance prediction, emoji prediction, and irony detection. We develop a unified framework, unified criteria for train/validation/test splits, and evaluate strong baselines inspired by current SotA in these tasks. We also evaluate transformer-based models, trained entirely and partially on Twitter data, with which we aim to establish a competitive high bar for subsequent contributions. The contributions of this paper are therefore as follows: (1) we compile, curate and release a suite of tasks under the umbrella of a new benchmark: \textbf{\textsc{TweetEval}}\footnote{The unified \textsc{TweetEval} benchmark is available at:\\\scriptsize \url{https://github.com/cardiffnlp/tweeteval}}, a unified framework comprising several tweet classification tasks; %CR effectively building a \textbf{unified framework} for this problem,
and (2) we \textbf{evaluate state-of-the-art LMs in this new framework}, and shed light on the effect of training with different corpora.

\begin{table*}[ht]
\renewcommand{\arraystretch}{1.18}
\resizebox{\textwidth}{!}{
\begin{tabular}{@{}lll@{}}
\toprule
\multicolumn{1}{l}{\large{\textbf{Dataset}}} & \multicolumn{1}{c}{\large{\textbf{Tweet}}}                                                                                                   & \multicolumn{1}{c}{\large{\textbf{Label}}} \\ \midrule
Emoji                       & Thx for showing this newbie passholder around @ Disneyland                                                                  & \includegraphics[height=0.45cm,width=0.45cm]{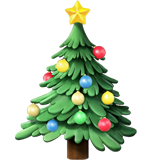}          \\
Emotion                     & I love swimming for the same reason I love meditating...the feeling of weightlessness.                                      & joy                       \\
Hate                        & Another illegal alien that shouldn't be in America killed an innocent American couple! \#BuildThatWall & hateful                   \\
Irony                       & Leaving whilst its dark is fun. \#not                                                                                        & ironic                    \\
Offensive                   & Are we all ready to sit and watch Indakurate Passcott play football?                                                        & non-offensive             \\
Sentiment                   & Hmmmmm where are the \#BlackLivesMatter when matters like this a rise... kids are a disgrace!!                               & negative                  \\
Stance\textit{(fem)}             & Rather be an ``ugly'' feminist then be these sad people that throws hat on people that believes in equality!            & in favour                \\ \bottomrule
\end{tabular}
}
\caption{Tweet samples for each of the tasks we consider in TweetEval, alongside their label in their original datasets. We use \textit{(fem)} to refer to the \textit{feminism} subset of the stance detection dataset.}
\label{tab:datasamples}
\end{table*}

\section{TweetEval: The Benchmark}
\label{tasks}

In this section, we describe the compilation, curation and unification procedure behind the construction of \textsc{TweetEval} and its corresponding tasks, as well as relevant statistics and evaluation metrics. We also show, in Table \ref{tab:datasamples}, a sample tweet and its corresponding label from the original task.

\subsection{Tasks}

%CR \footnote{All datasets included in TweetEval have a non-restrictive license for being shared.}
%Table \ref{table-data} includes the main statistics of all datasets included in TweetEval. As can be observed, there are tasks with diverging training data sizes (from a few hundreds instances for training in stance detection to over 40,000 instances in tasks such as emoji prediction and sentiment analysis).%, which also makes the benchmark more interesting from the machine learning point of view.

%\subsection{Tasks}
%\label{tasks}

\textbf{Emotion Recognition}.
This task consists of recognizing the emotion evoked by a tweet. We use the dataset of the most participated task of SemEval2018, ``Affects in Tweets'' \cite{mohammad2018semeval}. The original competition was framed as a multi-label classification problem, including 11 emotions. The integration into \textsc{TweetEval} consists of re-purposing this multi-label dataset into multi-class classification, keeping only the tweets labeled with a single emotion. Since the amount of tweets with single labels was scarce, we selected the most common four emotions
%\red{with relatively small overlap}
(Anger, Joy, Sadness, Optimism)\footnote{We selected those emotions with a minimum frequency of 300 examples in the training set.}. %The official evaluation metric is macro-averaged F1.%, as the labels are not balanced. 

\textbf{Emoji Prediction}.
%Given a tweet that includes one (and only one) emoji, the emoji prediction task requires predicting the emoji included in the tweet, using only its text content. 
This task consists in, given a tweet, predicting its most likely emoji, and is based on the Emoji Prediction challenge at Semeval2018 \cite{barbieri2018semeval}. It only considers tweets with one emoji (irrespective of its position), which is used as classification label. The test set is the same as in the original publication, but we limit the training and validation splits to 50,000 tweets, in order to comply with Twitter distribution policies. %\footnote{In order to comply with the Twitter policies, and be able to provide the text without the need for crawling the tweet with the ID.}. 
The label set comprises 20 different emoji, and due to their skewed distribution, this task proved to be highly difficult, with low overall numbers. Specifically, more than 42\% of the tweets are labeled with the 3 most frequent emoji
(\includegraphics[height=0.32cm,width=0.32cm]{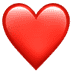}, \includegraphics[height=0.32cm,width=0.32cm]{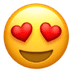}, and \includegraphics[height=0.32cm,width=0.32cm]{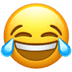}). %The official metric for this task is macro-averaged F1.

\textbf{Irony Detection}. This task consists of recognizing whether a tweet includes ironic intents or not. We use the Subtask A dataset of the SemEval2018 Irony Detection challenge \cite{van2018semeval}. Note that this dataset was artificially balanced to make the task more accessible. %\luis{don't like easier}.%o ease this very challenging problem, the organizer balanced the dataset. 
 %Irony detection is a very challenging problem, and the amount of ironic tweet is very little in a real scenario. To ease the task, the organizer balanced the dataset. 
%Moreover, because there was no validation set provided we extracted it from the training set. %The official evaluation metric is F1 on the \textit{irony} class. 

\textbf{Hate Speech Detection}. This task consists in predicting whether a tweet is hateful or not against any of two target communities: immigrants and women. Our dataset of choice stems from the SemEval2019 Hateval challenge \cite{basile-etal-2019-semeval}.%, where the official metric was macro-averaged F1. %\luis{accuracy no? official to rank systems was macro F1}.

\textbf{Offensive Language Identification}. This task consists in identifying whether some form of offensive language is present in a tweet. For our benchmark we rely on the SemEval2019 OffensEval dataset \cite{zampieri-etal-2019-semeval}. %This dataset only included training and test splits and thus we create the development split accordingly from the training set. %\fre{The official metric was F1-macro.}
%\luis{I think at the beginning of sect 2 we should say how we create the dev set in cases where it was not provided, and discuss the most frequent evaluation metric, and say that these two things apply to each dataset unless otherwise noted, because it is very redundant to read this stuff for each task}.% in order to include it into our benchmark, we used 10\% of the training set as validation.

\begin{table}
\begin{center}
{
\setlength{\tabcolsep}{4.8pt}
\scalebox{0.93}{ 
\begin{tabular}{lcrrr}

\toprule
{\bf Task}	& {\bf Lab}	& {\bf Train}  & {\bf Val}   & {\bf Test}     \\
\midrule

Emoji prediction	& 20	& 45,000  & 5,000   & 50,000     \\
Emotion	rec. & 4	& 3257  & 374   & 1421     \\%\hline
Hate speech det.	& 2	& 9,000  & 1,000   & 2,970     \\
Irony detection	& 2	& 2,862  & 955   & 784     \\
Offensive lg. id.	& 2	& 11,916  & 1,324   & 860     \\
Sent. analysis	& 3	& 45,389  & 2,000   & 11,906     \\
Stance detection & 3 & 2620 & 294 & 1249 \\
\midrule
\midrule
Stance/Abortion	& 3	& 587  & 66 & 280     \\
Stance/Atheism	& 3	& 461  & 52   & 220     \\
Stance/Climate	& 3	& 355  & 40   & 169     \\
Stance/Feminism	& 3	& 597  & 67   & 285     \\
Stance/H. Clinton	& 3	& 620  & 69   & 295     \\

%Stance (Abortion)	& 3	& 587  & 66 & 280     \\
%Stance (Atheism)	& 3	& 461  & 52   & 220     \\
%Stance (Climate)	& 3	& 355  & 40   & 169     \\
%Stance (Feminism)	& 3	& 597  & 67   & 285     \\
%Stance (Hillar C)	& 3	& 620  & 69   & 295     \\

\bottomrule

\end{tabular}
}
}
\end{center}
\caption{\label{table-data} Number of labels and instances in training, validation, and test sets for each dataset. The specific statistics of each target domain in the stance detection task is included at the bottom.}
\end{table}

\textbf{Sentiment Analysis}. %Sentiment Analysis is one of the most investigated tasks in NLP. 
The goal for the sentiment analysis task is to recognize if a tweet is positive, negative or neutral. 
We use the Semeval2017 dataset for Subtask A \cite{rosenthal2019semeval}, which includes data from previous runs (2013, 2014, 2015, and 2016) of the same SemEval task. %The official evaluation metric is macro-averaged recall.

\textbf{Stance Detection}. Stance detection is the task to determine, given a piece of text, whether the author has a favourable, neutral, or negative position towards a proposition or target. We use the SemEval2016 shared task on Detecting Stance in Tweets \cite{mohammad2016semeval}. In the original task, five target domains are given: abortion, atheism, climate change, feminism and Hillary Clinton. Unlike the other tasks, %, in stance detection,
training is provided separately for each target domain, which we use to extract individual validation sets.
%\footnote{The appendix includes stance detection dataset statistics by target domain. %\red{TODO: Add now these stats in the main paper?}
%} 
%We detach the validation set from the train set using our unified procedure. 
%The official metric is average of the F1 score of the ``against'' and ``in favor'' classes over all domains.

\subsection{Statistics and evaluation metrics}

Table \ref{table-data} includes the \textsc{TweetEval} datasets statistics after unification.\footnote{The validation sets are randomly sampled from the training set for those tasks where no validation split is provided in the original dataset.} Data sizes range from a few hundred instances for training to over 40,000. Note that the preprocessing pipeline is equal for all tasks: user mentions are anonymized %CR \footnote{User mentions are replaced with the token \textit{@user}.} 
and line breaks and website links are removed. %In what follows we flesh out the tasks available in \textsc{TweetEval}.

 %(replacing the mentions with the token @user), remove both break lines and internet links.
\paragraph{Evaluation metrics.}
We use the same evaluation metric from the original tasks, which is macro-averaged F1 over all classes, in most cases. There are three exceptions: stance (macro-averaged of F1 of favor and against classes), irony (F1 of ironic class), and sentiment analysis (macro-averaged recall). Similar to GLUE \cite{wang2019glue}, we also introduce a global metric (TE) based on the average of all dataset-specific metrics.

%\subsection{Statistics}
%Table \ref{table-data} includes the main statistics of all datasets included in TweetEval. As can be observed, there are tasks with diverging training data sizes (from a few hundreds instances for training in stance detection to over 40,000 instances in tasks such as emoji prediction and sentiment analysis).%, which also makes the benchmark more interesting from the machine learning point of view.

% \multirow{nrows}[bigstruts]{width}[fixup]{text}

\section{Language Models for Tweet Classification}
\label{method}
%Language models (very brief), explain techniques (from scratch, base and then train on Twitter, etc.). 
%\red{TOFINISH: common technique: train a masked language model, then fine-tune it on a classification task.}

%Given a tweet, the task of tweet classification consists of associating it with a label from a pre-defined set. For example, in a simplified sentiment analysis setting the pre-defined labels could be positive, negative and neutral. %In the following we describe some traditional linear models and then explain recent techniques based on language models.

Transformer-based LMs such as GPT \cite{radford2018improving}, BERT \cite{devlin-etal-2019-bert} or XLNET \cite{yang2019xlnet} have taken the NLP field by storm, outperforming previous linear models and neural network methods based on LSTMs or CNNs in many tasks, including sentence and text classification \cite{wang2019glue}. 
%In our setting, the task of tweet classification consists of associating a given tweet with a label from a pre-defined set. For example, in our simplified sentiment analysis setting, the pre-defined labels are positive, negative, and neutral.

The functioning of these language models for tweet classification is conceptually simple. First, they are trained on a large unlabeled corpus. Then, they are fine-tuned to the task for where an appropriate training set exists. For social media text, however, one may question whether existing pre-trained models trained on standard corpora %CR such as Wikipedia or BooksCorpus \cite{zhu2015aligning} %(as in the case of BERT, for example) 
are optimal. We thus compare three different strategies which differ in the training data:
%\begin{enumerate}
    %\item 
(1) Using an existing large pre-trained LM;
    %\item 
(2) using an existing architecture, but training from scratch using only Twitter data; and
    %\item the same architecture as an existing pre-trained LM
(3) starting with an original pre-trained LM and continue to train with Twitter data, keeping the original tokenizer and the same masked LM loss. 
 
%\end{enumerate}
%(3GB)

%\red{TODO: Rewrite, make it more general here (just explaining the techniques) and explain the specifics (Roberta) in experimental setting}

%For our experiments we use three versions of roberta-base.
We consider these three techniques as we are interested in exploring whether a Twitter-specific LM should be trained on Twitter only 
%(as in the concurrent work of \newcite{BERTweet}) 
or if it should be initialized with weights learned during pre-training on standard corpora, \textit{and then} be trained on Twitter.
%start with original data convenient to use a LM already trained on more standard text
The latter option has indeed three theoretical advantages: (1) these models are generally trained on large amounts of text corpora, and reproducing the same experiment would be extremely expensive even if we had same amount of Twitter data; (2) learning on different types of text corpora make the models more robust and knowledgeable about the world; and (3) some models such as RoBERTa \cite{liu2019roberta} or GPT-2 \cite{radford2019language} are not unfamiliar with internet language and slang, as part of their underlying training corpora contains Reddit data (38GB).%\luis{I think we should also say that learning on general text is good bc the model learns stuff about the world, and generic grammar and syntax which are going to be useful anyway because the underlying language is the same, what we want is actually a model that speaks a language first, and that can adapt to slang/jargon afterwards.}

%This employs starting with a random parameters initialization, and re-traing the Byte-Pair Encoding (BPE) tokenizer \cite{sennrich2015neural} from scratch on the Twitter corpus.

%\red{TO FINISH: want to say 1) roberta model was trained on a huge amount of text, and reproducing the same experiment would be extremely expensive even if we had same amount of Twitter data, hence better to use this model as a starting point (2) Roberta model is not unfamiliar with internet language and slang, as it was trained also on a corpus of Reddit data (38GB) \cite{radford2019language}}

\section{Evaluation}

%In this section, we perform a comparative analysis on all TweetEval tasks.

\begin{table*} %[bp]
\renewcommand{\arraystretch}{1.25}
\setlength{\tabcolsep}{2.0pt}
\scalebox{0.75}{ 

\begin{tabular}{cc|c|c|c|c|c|c|c||c}
\toprule
\textbf{} &
  \textbf{} &
  \textbf{Emoji} &
  \textbf{Emotion} &
  \textbf{Hate} &
  \textbf{Irony} &
  \textbf{Offensive} &
  \textbf{Sentiment} &
  \textbf{Stance} &
  \textbf{ALL} \\ \hline \hline
\multicolumn{1}{c|}{\multirow{6}{*}{Val}} &
  SVM &
  25.0 &
  63.8 &
  73.1 &
  63.4 &
  72.7 &
  68.4 &
  67.9 &
  62.0 \\
\multicolumn{1}{c|}{} &
  FastText &
  23.2 &
  62.9 &
  71.7 &
  62.7 &
  70.0 &
  62.2 &
  67.3 &
  60.0 \\
\multicolumn{1}{c|}{} &
  BLSTM &
  19.4 &
  62.6 &
  72.1 &
  60.6 &
  72.1 &
  61.9 &
  63.4 &
  58.9 \\
\multicolumn{1}{c|}{} &
  RoB-Bs &
  24.7{\small $\pm$0.3} (24.3) &
  73.1{\small $\pm$1.7} (74.9) &
  76.5{\small $\pm$0.3} (76.6) &
  73.7{\small $\pm$0.6} (73.7) &
  77.1{\small $\pm$0.6} (77.6) &
  71.4{\small $\pm$1.9} (72.7) &
  71.4{\small $\pm$1.9} (73.9) &
  67.7 \\
\multicolumn{1}{c|}{} &
  RoB-RT &
  24.4{\small $\pm$1.5} (\textbf{26.2}) &
  75.4{\small $\pm$1.5} (\textbf{77.0}) &
  77.8{\small $\pm$1.1} (\textbf{79.6}) &
  74.7{\small $\pm$1.5} (\textbf{75.6}) &
  77.2{\small $\pm$0.6} (\textbf{77.7}) &
  73.0{\small $\pm$1.2} (\textbf{74.2}) &
  72.9{\small $\pm$1.0} (\textbf{75.2}) &
  \textbf{69.4} \\
\multicolumn{1}{c|}{} &
  RoB-Tw &
  23.4{\small $\pm$1.1} (24.6) &
  67.6{\small $\pm$0.9} (68.6) &
  74.3{\small $\pm$2.0} (76.6) &
  70.0{\small $\pm$0.3} (70.7) &
  76.1{\small $\pm$0.6} (76.2) &
  70.5{\small $\pm$1.0} (69.4) &
  68.3{\small $\pm$2.4} (71.4) &
  65.4 \\ \hline
\multicolumn{1}{c|}{\multirow{6}{*}{Test}} &
  SVM &
  29.3 &
  64.7 &
  36.7 &
  61.7 &
  52.3 &
  62.9 &
  67.3 &
  53.5 \\
\multicolumn{1}{c|}{} &
  FastText &
  25.8 &
  65.2 &
  50.6 &
  63.1 &
  73.4 &
  62.9 &
  65.4 &
 58.1 \\
\multicolumn{1}{c|}{} &
							
  BLSTM &
  24.7 &
  66.0 &
  52.6 &
  62.8 &
  71.7 &
  58.3 &
  59.4 &
  56.5  \\
\multicolumn{1}{c|}{} &
  RoB-Bs &
  30.9{\small $\pm$0.2} (30.8) &
  76.1{\small $\pm$0.5} (76.6) &
  46.6{\small $\pm$2.5} (44.9) &
  59.7{\small $\pm$5.0} (55.2) &
  79.5{\small $\pm$0.7} (78.7) &
  71.3{\small $\pm$1.1} (72.0) &
  68{\small $\pm$0.8} (70.9) &
  61.3 \\
\multicolumn{1}{c|}{} &
  RoB-RT &
  31.4{\small $\pm$0.4} (\textbf{31.6}) &
  78.5{\small $\pm$1.2} (\textbf{79.8}) &
  52.3{\small $\pm$0.2} (\textbf{55.5}) &
  61.7{\small $\pm$0.6} (62.5) &
  80.5{\small $\pm$1.4} (\textbf{81.6}) &
  72.6{\small $\pm$0.4} (\textbf{72.9}) &
  69.3{\small $\pm$1.1} (\textbf{72.6}) &
  \textbf{65.2} \\
\multicolumn{1}{c|}{} &
  RoB-Tw &
  29.3{\small $\pm$0.4} (29.5) &
  72.0{\small $\pm$0.9} (71.7) &
  46.9{\small $\pm$2.9} (45.1) &
  65.4{\small $\pm$3.1} (\textbf{65.1}) &
  77.1{\small $\pm$1.3} (78.6) &
  69.1{\small $\pm$1.2} (69.3) &
  66.7{\small $\pm$1.0} (67.9) &
  61.0 \\ \cline{2-10} 
\multicolumn{1}{c|}{} &
  \textit{SotA} &
  36.0* &
  - &
  65.1 &
  70.5 &
  82.9 &
  68.5 &
  71.0 &
  - \\ \hline \hline
\multicolumn{2}{c|}{\textbf{Metric}} &
  M-F1 &
  M-F1 &
  M-F1 &
  F$^{(i)}$ &
  M-F1 &
  M-Rec &
  AVG (F$^{(a)}$,$F^{(f)}$) & TE
   \\ \bottomrule
\end{tabular}
}
\caption{\label{table-results} TweetEval validation and test results. For neural models we report both the average result from three runs and its standard deviation, and the best result according to the validation set (parentheses). \textit{SotA} results correspond to the best systems in the original shared tasks - they are included for completeness as they not directly comparable.  %as were reported under different settings and not in this unified benchmark, . 
Splits might differ, and * indicates that a larger training set is used.}
\end{table*}

\subsection{Experimental setting}

%In the following we describe the common experimental setting.
\textbf{Neural language model}.
%For our experiments we make use of RoBERTa \cite{liu2019roberta} in its base version. 
Among all the available language models we selected RoBERTa \cite{liu2019roberta} as it is one of the top performing systems in GLUE. Moreover, it does not employ the Next Sentence Prediction (NSP) loss \cite{devlin2018bert}, making the model more suitable for Twitter where most tweets are composed of a single sentence. 

\noindent \textbf{Language model pre-training}. %As explained in Section \ref{method},
We use three different RoBERTa variants: 
%of the reference language model, in this case RoBERTa:
pre-trained RoBERTa-base\footnote{RoBERTa-base was trained on 160G of uncompresed text.} (RoB-Bs), the same model but re-trained on Twitter (RoB-RT) and trained on Twitter from scratch (RoB-Tw). %\red{For the pre-trained models we considered the base version, trained on 160G of uncompresed text.} %For the three version of RoBERTa-base that we use the first one do not need additional finetuning as it is used as it is. The second one and the third one use the same language model loss function of the original roberta-base model. 
RoB-RT and RoB-Tw are trained with early stopping on the validation split and learning rate $1.0e^{-5}$. %\footnote{\url{github.com}}. 
Both models converged after about 8/9 days on 8 NVIDIA V100 GPUs.\footnote{We used the Huggingface \textit{transformers} library %\footnote{\url{https://huggingface.co/transformers/}}
. The estimated cost for each language model is USD 4,000 on Google Cloud.}

\noindent \textbf{Twitter corpus}. We train RoB-RT and RoB-Tw on %In order to train the two later variants of the language model 
%and the FastText embeddings 
%on Twitter data, we used a corpus of tweets %\footnote{\url{https://archive.org/}}
%that was crawled with the stream API from May 2018 to August 2019. From this corpus, we selected tweets in English (using the automatic labeling provided by Twitter), tweets with at least three tokens, and tweets that did not include any web links (to avoid bot tweets and spam advertising). After filtering we obtained a corpus of 60 millions tweets, and 584 millions tokens (3.6G uncompressed text).
%we used a corpus of 
60M tweets\footnote{584 million tokens (3.6G of uncompressed text).} obtained by extracting a large corpus of English tweets\footnote{Crawled with the stream API from May'18 to August'19.} (using the automatic labeling provided by Twitter). We only considered tweets with at least three tokens and without URLs, as to avoid bot tweets and spam advertising.

\noindent \textbf{Classification fine-tuning}. We use the same classification fine-tuning method used in \newcite{liu2019roberta}: we add one dense layer to reduce the dimensions of the RoBERTa's last layer to the number of labels in the classification task, and fine-tune the model on each classification task, training all the parameters simultaneously. %CR (i.e., without employing any incremental freezing method \cite{sun2019fine}). 
We run a minimum parameter search on the starting learning rate ($1.0e^{-3}$, $1.0e^{-4}$, $1.0e^{-5}$, and $1.0e^{-6}$), use early stopping (5 epochs) on the validation set and run each experiment three times with different seeds ($1$,$2$,$3$). Then, we select the highest performing learning rate on the validation set, and use the corresponding model to evaluate on the test set. % (along with the average and standard deviation of the three runs with different seeds).
%reporting both the average results of the three runs, and the best result (picked using the result on validation set).

\noindent \textbf{Baselines}. %Linear models such as SVMs \cite{} or logistic regression \cite{} coupled with frequency-based hand-crafted features have been traditionally used to classify tweets \cite{}(\red{Add citations}). 
%These models have slowly been replaced by more powerful (although more time-consuming and data inefficient) language models, which we are going to describe in the following section. However, recently there has been a revamped of these traditional methods by including character ngrams or subword units, which are features very well suited to the informal nature of social media text. 
%Linear models have been traditionally a strong baseline for tweet classification. 
FastText \cite{joulin-etal-2017-bag} provides an efficient baseline based on standard features and subword units. %\footnote{\url{https://fasttext.cc}} 
We also include an SVM-based baseline with both word and character n-gram features, a model and feature set that has %even been able to outperform neural methods in 
seen great success in recent Twitter-based shared tasks such as emoji prediction \cite{ccoltekin2018tubingen} and stance prediction \cite{mohammad2018semeval}. %outperforming data-hungry neural models such as LSTMs and CNNs.
%\footnote{For the SVM baseline zwe used the code from \newcite{ccoltekin2tubingen} available at \red{XXXX}.} 
We finally report the results of a bi-directional LSTM%\cite{hochreiter1997long}
.\footnote{The LSTM has 128 cells, an embedding layer of 100 dimensions, dropout (0.5) and, similarly to the language models, the four learning rate values are tuned in the validation set.} Both FastText and the LSTM use 100-dimensional FastText word embeddings \cite{bojanowski-etal-2017-enriching} trained on the 60M Twitter corpus for the lookup table initialization.

%\noindent \emph{\textbf{Evaluation metrics}}. We use all the official metrics from the original tasks (see Section \ref{tasks}). Similar to GLUE \cite{wang2019glue}, we introduce a global metric (TE) based on the average of all dataset-specific metrics. \luis{If we say this at the beginning of sect. 2, alongside the official TE metric, this paragraph could go.}

%%%%%%%%%%%%%%%%%%%%%%%%%%%%%%%%%%%%%%%%%%%%%%%%%%%%%%%%%%%%%%%%%%%%%%%%%%%%%

%%%%%%%%%%%%%%%%%%%%%%%%%%%%%%%%%%%%%%%%%%%%%%%%%%%%%%%%%%%%%%%%%%%%%%%%%%%%%

\subsection{Results}

%red{Add table with results (mean average + std deviation + best/worst result?). We can also add the dev results (here or in the sup material)}

Table \ref{table-results} shows the results of all comparison systems on \textsc{TweetEval}. Perhaps surprisingly, RoBERTa-Base (RoB-Bs) performs well on all tasks, even outperforming the model trained on Twitter data only (RoB-Tw) in most tasks. This can also be attributed to the fact that Twitter is not only noisy text, and formal text can be also found regularly \cite{hu2013dude,xu2017shakespeare}. %\luis{precisely}
Using more Twitter data for training might further improve the results of RoB-Tw, but this would also translate into an even more expensive training.
%More Twitter data might be necessary to outweigh its current grasp of the standard language present in Twitter.
%but also that more Twitter data might be necessary to outweigh its current grasp of the standard language present in Twitter.%ouperform it.
However, %the best model overall is 
RoBERTa-Base coupled with additional training on the same Twitter corpus (i.e. RoB-RT) proves more effective.

The only task where a model trained from scratch on Twitter performs better is Irony detection, where RoB-Tw shows to better generalize (RoB-RT F1 drops 13 points from validation to test set, while Rob-Tw F1 5 points).  This can be due to two factors: (1) irony used on social media might differ from irony on standard text, (2) tweets in our training data are generally short (79.3 characters on average compared to over 100 characters for most other tasks), and therefore tokenizing the text in less word pieces, and potentially less OOVs, becomes more important to generalize.
%\footnote{We provide a per-dataset comparison of the effects of (word piece) tokenization in the supplementary material.}
%In the supplementary material there is a comparative of tokenizer on all datasets.}.%CR and (3) It also depends on the metric employed, if we look at the macro-averaged F1, the results for RoB-Bs, RoB-RT, and RoB-Tw are 65.8, 70.9, and 69.6.
We note that the low results in the task of emoji prediction (when compared to those obtained in the official SemEval task) are due to the downscaling of the training data. Because of Twitter's data distribution policy, at \textsc{TweetEval} we release at most 50k tweets per task, whereas in the original competition, by id sharing, the training data was one order of magnitude bigger. As for the results in the hate speech task, the difference in performance between validation and test set is mainly due to these splits being collected at different timespans, as pointed out by the organizers of the task \cite{basile-etal-2019-semeval}. This causes a disparity in topic distribution and thus low performance of the systems optimized towards the validation set.

%\red{Explain low overall results for hate speech and emoji prediction (different distribution, tweets from different time, etc.)}

%\subsection{Analysis}

%Analysis

%ite this paper on biases on datasets (Valerio) \cite{wiegand-etal-2019-detection}

\subsection{Tokenizer analysis}

%The main difference between RoB-Bs, RoB-RT and RoB-Tw is the tokenizer, as RoB-Bs and RoB-RT include the original RoBERTa-base tokenizer, while RoB-Tw includes a new BPE tokenizer retrained on our corpus of tweets. 
Table \ref{table-tokenizer} includes number of tokens\footnote{Tokenized with the Twitter-specific ``Twikenizer'': \url{github.com/Guilherme-Routar/Twikenizer}} per tweet for each of the tasks and the difference between word pieces of the pre-trained RoBERTa-base and RoBERTa trained on Twitter from scratch.
%which apply the Byte-Pair Encoding (BPE) tokenizer \cite{sennrich2015neural}.
This comparison is useful to understand if a model recognizes more or less tokens: if the difference between the two RoBERTa tokenizers is high, it means that one model had to split more times a word.
We can note that the biggest difference in wordpieces between RoB-Bs and Rob-Tw is 6.8\% in the hate detection task. This is expected as these tweets include less standard words, such as insults. On the other hand, except for perhaps emotion recognition and offensive language identification, the difference is not significant, considering that the original RoBERTa tokenizer was not trained on Twitter text. %In other words the advantage of Rob-Tw with a novel tokenizer trained on Twitter, does not seem to be very advantageous, as only a small percentage of words are better recognized. 
Moreover, even if the tokenizer of Rob-RT was not retrained from scratch, this does not mean that Rob-RT could not learn new tokens as they could be learned as sequence of characters during the language modeling re-training phase. This is also the case of emoji, which were not learned in the original RoBERTa model, but BTE includes all their Unicode bytes. %, Rob-RT learned emoji semantics during the unsupervised phase as sequence of bytes.

\begin{table}[ht]
\renewcommand{\arraystretch}{1.25}
\setlength{\tabcolsep}{3.0pt}
\scalebox{0.85}{ 
\begin{tabular}{lc|ccc}
\toprule
\textbf{Task} & \textbf{Tokens} & \textbf{RoB-Bs} & \textbf{RoB-Tw} & \textbf{\% Diff} \\ \midrule
Emoji & 14.3 {\small $\pm$7.4} & 22.4 {\small $\pm$7.4} & 21.6 {\small $\pm$6.8} & 2.8 {\small $\pm$6.9}  \\
Emotion & 19.2 {\small $\pm$10.2} & 27.2 {\small $\pm$10.2} & 25.7 {\small $\pm$9.6} & 5.1 {\small $\pm$8.1}  \\
Hate & 25.6 {\small $\pm$19.7} & 38.6 {\small $\pm$19.7} & 36 {\small $\pm$18.9} & 6.8 {\small $\pm$8.2}  \\
Irony & 17.9 {\small $\pm$9.3} & 26.1 {\small $\pm$9.3} & 25.1 {\small $\pm$8.9} & 3.8 {\small $\pm$7.1}  \\
Sentiment & 18.9 {\small $\pm$9.2} & 26.7 {\small $\pm$9.2} & 26.2 {\small $\pm$9.1} & 1.4 {\small $\pm$8.5}  \\
Offensive & 28.4 {\small $\pm$20.9} & 41.9 {\small $\pm$20.9} & 39.4 {\small $\pm$19.7} & 5.7 {\small $\pm$8.5}  \\
Stance & 20.6 {\small $\pm$7.1} & 30.7 {\small $\pm$7.1} & 30.5 {\small $\pm$6.9} & 0.5 {\small $\pm$4.8}  \\

\bottomrule
\end{tabular}
}
\caption{\label{table-tokenizer} Tokenization statistics for all \textsc{TweetEval} tasks.
``Tokens'' is the average number of tokens in each tweet using Twikenizer. RoB-RT and Rob-Tw refers to the average number of word pieces after tokenization with the original Roberta-base and with the model trained from scratch. ``Diff'' is the relative difference (\%) of tokens in each tweet between these two tokenizers (if the difference is positive, the original RoBERTa includes more tokens). For stance detection, we computed the average statistics among the five targets.}

\end{table}

\section{Conclusion}

We have presented \textsc{TweetEval}, a unified benchmark for tweet classification consisting of seven heterogeneous tasks that are core to social media NLP research. Along with the benchmark, we have included strong baselines as reference, and ran an analysis of LMs with different training strategies. Our results suggest that using a pre-trained LM may be sufficient, but can improve if topped with extra-training on in-domain data.%$ (i.e. Twitter).

For this initial benchmark and in the interest of reproducibility and accessibility, we focused on a fixed setting (i.e. classification). However, we acknowledge that other important tasks may need to be evaluated differently. Thus,
%it is clear that are other important tasks that need to be evaluated differently. 
for future work we would like to include more tasks in the context of social media NLP research. Potential improvements include, for example, accounting for the original multi-label nature of emotion classification, or covering more than only 20 emoji in emoji prediction. %For example, emotion classification was initially multi-label, with more than four labels, or emoji prediction should consist of more than only twenty emoji. 
There are also other scenarios to be addressed as well, like sequence tagging \citep{baldwin2015shared, postwitter}, multimodality \citep{2016arXiv160802289S,lu2018visual}, and code-switching tasks \citep{barman2014code,vilares2016cs}. This is similar to the evolution of GLUE \cite{wang2019glue} into SuperGLUE \cite{wang2019superglue}, with both benchmarks contributing to the development of the field in different ways. It is also important to highlight that these datasets do not represent their underlying tasks as a whole but only a subsample,  %and that they have been gathered in specific ways\luis{?????},
and therefore contain biases - automatic models trained on them might not be able to generalize to other specific settings %and automatic models trained on them are not always generalizable to more general settings 
\cite{augenstein2017generalisation,wiegand-etal-2019-detection}.

Finally, this benchmark could foster research in multitask learning. The fact that several similar tasks co-exist (e.g. sentiment analysis and emotion recognition, or hate speech detection and offensive language identification) can lead to interesting analyses where the similarity of these tasks is exploited. %based on the similarity of the tasks. 
%Likewise, developing benchmarks for languages other than English could help investigate different phenomena across languages and enable the testing of cross-lingual NLP technologies, similar to the recent XTREME multilingual benchmark for non-social media text \cite{hu2020xtreme}.

%\end{spacing}

%%%%%%%%%%%%%%%%%%%%%%%%%%%%%%%%%%%%%%%%%%%%%%%%

%DIFFERENT TRAIN-TEST DISTRIBUTION: \cite{augenstein2017generalisation}

%\section*{Acknowledgments}

%The acknowledgments should go immediately before the references. Do not number the acknowledgments section.
%Do not include this section when submitting your paper for review.

\bibliographystyle{acl_natbib}
\bibliography{anthology,emnlp2020}

\end{document}

\maketitle
\begin{abstract}
Supplementary material of the EMNLP submission \textit{TweetEval: Unified Benchmark and Comparative Evaluation for Tweet Classification}

\end{abstract}

\section{Stance detection dataset}

Table \ref{table-stance} includes the statistics for the five target domains available in the stance detection dataset \cite{mohammad2016semeval}. We report this statistics as, even if the final metric is the average between F1 of the favour class and F1 of the against class across all the targets, we train on each single task as done by the majority of systems in the Semeval 2016 shared task.

% \multirow{nrows}[bigstruts]{width}[fixup]{text}
\begin{table}[h]
\begin{center}
{
\setlength{\tabcolsep}{4.8pt}
\scalebox{0.93}{ 
\begin{tabular}{|l|c|rrr|r}
%\toprule
\cline{2-5}
%\multicolumn{1}{l|}{Task} & \multicolumn{1}{r|}{\bf Lab} & \multicolumn{1}{r}{\bf Train} & \multicolumn{1}{r}{\bf Dev} & \multicolumn{1}{r|}{\bf Test} \\
%\bf Eval. \\
\hline
{\bf Target}	& {\bf Lab}	& {\bf Train}  & {\bf Dev}   & {\bf Test}     \\
\hline
\hline
Abortion	& 3	& 587  & 66 & 280     \\
Atheism	& 3	& 461  & 52   & 220     \\
Climate	& 3	& 355  & 40   & 169     \\
Feminism	& 3	& 597  & 67   & 285     \\
Hillary Clinton	& 3	& 620  & 69   & 295     \\
\hline
\hline
ALL & 3 & 2620 & 294 & 1249 \\
\hline

\hline

\end{tabular}
}
}
\end{center}
\caption{\label{table-stance} Number of labels and instances in training, development and test sets split by target word of the stance detection dataset.}
\end{table}

\section{Number of tokens per datasets}

Table \ref{table-tokenizer} includes number of chars per tweet for each of the tasks and the difference between word pieces of the pre-trained RoBERTa-base and RoBERTa trained from scratch, which apply the Byte-Pair Encoding (BPE) tokenizer \cite{sennrich2015neural}.
This is useful to understand if a model recognizes more or less tokens: if the number is high, it means that the model had to split more times a word.

\begin{table}[h]
\renewcommand{\arraystretch}{1.25}
\setlength{\tabcolsep}{4.5pt}
\scalebox{0.95}{ 
\begin{tabular}{lcccc}
\hline
\multirow{2}{*}{\textbf{Task}} & \multicolumn{2}{c}{\textbf{Train}}     & \multicolumn{2}{c}{\textbf{Test}}      \\
                               & \textbf{Chars} & \textbf{Diff} & \textbf{Chars} & \textbf{Diff} \\ \hline
Emoji     & 72.02  & 0.69{\small $\pm$1.69} & 71.3   & 0.77{\small $\pm$1.88} \\
Emotion   & 92.37  & 0.99{\small $\pm$2.74} & 90.89  & 1.45{\small $\pm$3.1}  \\
Hate      & 123.86 & 0.93{\small $\pm$3.14} & 132.57 & 2.58{\small $\pm$3.49} \\
Irony     & 79.25  & 1.12{\small $\pm$2.17} & 85.13  & 1.02{\small $\pm$2.1}  \\
Sentiment & 108.01 & 0.29{\small $\pm$1.68} & 95.64  & 0.43{\small $\pm$2.5}  \\
Offensive & 125.76 & 1.64{\small $\pm$3.97} & 144.74 & 2.52{\small $\pm$4.42} \\
Stance    & 106.03 & 0.3{\small $\pm$1.61}  & 107.47 & 0.21{\small $\pm$2.25} \\
\hline
\end{tabular}
}
\caption{\label{table-tokenizer} Statistics from the seven tasks from TweetEval. %the roberta-base tokenizer and new trained tokenizer from scratch on Twitter data.
``Chars'' is the average number of characters in each tweet, and ``Diff'' is the mean of the difference of number of tokens in each tweet between RoBERTa-base and RoBERTa-base from scratch (if the difference is positive, the original RoBERTa includes more tokens.)}

\end{table}

\section{TweetEval: Development results}

Table \ref{table-results} includes all both development and test results of all comparison system in TweetEval.

\bibliographystyle{acl_natbib}
\bibliography{anthology,emnlp2020}